\documentclass[sigconf,nonacm]{acmart}
\usepackage{graphicx}
\usepackage{subfig}
\usepackage{float}
\usepackage{afterpage}
\usepackage{adjustbox}
\usepackage{booktabs}
\usepackage{multirow}
\usepackage{tcolorbox}
\tcbuselibrary{breakable}
\usepackage{graphicx}
\usepackage{subfig}
\usepackage{xcolor} % 색상 사용 시 반드시 필요
\usepackage{float}
\usepackage{listings}

\begin{document}

% an optional teaser image, which appears on the first page of the paper / article, below the title / author information and above
% the content of the paper / article.

% author information.
\author{Hyunseok Park}
\affiliation{%
  \institution{HDC LABS}
  \country{Republic of Korea}
}
\email{hyunskki@hdc-labs.com}

\author{Jihyeon Kim}
\affiliation{%
  \institution{HDC LABS}
  \country{Republic of Korea}
}
\email{jh991219@hdc-labs.com}

\author{Jongeun Kim}
\affiliation{%
  \institution{HDC LABS}
  \country{Republic of Korea}
}
\email{JongeunKim@hdc-labs.com}

\author{Dongsik Yoon}
\authornote{Corresponding author.}
\affiliation{%
  \institution{HDC LABS}
  \country{Republic of Korea}
}
\email{kevinds1106@hdc-labs.com}
%\author{Anonymous Author}

% This command provides customization of the author information in page headers. 
% It can safely be commented out, and a default author list generated for this purpose.
% A lengthy author list can overlap with the title information in the page headers.
%\renewcommand{\shortauthors}{Smith, et al.}

% Title of the work. 
% If your title is lengthy, consider using the "short title" parameter:
%  \title[short title]{full title}

\title{CHOP: Chunkwise Context-Preserving Framework \\ for RAG on Multi Documents}

% CCS Concepts. Required for any ACM work over two pages in length.
% Visit http://dl.acm.org/ccs and generate this content for your work.

% Keywords. Optional.


% Abstract.

\begin{abstract}
Retrieval-Augmented Generation (RAG) systems lose retrieval accuracy when similar documents coexist in the vector database, causing unnecessary information, hallucinations, and factual errors. To alleviate this issue, we propose CHOP, a framework that iteratively evaluates chunk relevance with Large Language Models (LLMs) and progressively reconstructs documents by determining their association with specific topics or query types. CHOP integrates two key components: the CNM-Extractor, which generates compact per-chunk signatures capturing categories, key nouns, and model names, and the Continuity Decision Module, which preserves contextual coherence by deciding whether consecutive chunks belong to the same document flow. By prefixing each chunk with context-aware metadata, CHOP reduces semantic conflicts among similar documents and enhances retriever discrimination. Experiments on benchmark datasets show that CHOP alleviates retrieval confusion and provides a scalable approach for building high-quality knowledge bases, achieving a Top-1 Hit Rate of 90.77\% and notable gains in ranking quality metrics.
\end{abstract}

% Processes the title / author / affiliation information and builds the first part of the generated PDF.

\maketitle
\section{Introduction}
%up-to-date information = fresh information
%broadening real-world = improving applicability to real-world scenarios
%Even when a relevant -> Even when a relevant document is retrieved, absent local context can render evidence uninterpretable, degrading recall and answer robustness. Moreover, irrelevant or weakly relevant material in retrieved context can mislead LLMs.

Large Language Models \textbf{(LLMs)} have demonstrated remarkable performance across a wide range of tasks. Nonetheless, they often underperform in knowledge-intensive or domain-specific scenarios, particularly when queries extend beyond their training distribution or require up-to-date information, leading to hallucinations~\cite{llm1, llm2}. Retrieval-Augmented Generation \textbf{(RAG)} offers a promising solution by retrieving semantically relevant evidence from external knowledge sources, grounding model outputs to mitigate factual errors and enhance practical utility~\cite{lewis2020retrieval, naiverag}. However, a major limitation of current RAG pipelines arises from length-based segmentation, which disrupts continuity by fragmenting coreference and local references (e.g., “this method,” “Eq. (3)”), leading reducing interpretability and recall. In multi-document contexts, redundancy, duplication, and conflicting references further increase retrieval ambiguity and destabilize grounding. Moreover, irrelevant or weakly related passages can mislead LLMs, compounding the risk of inaccurate outputs.

Complementary to these observations, \textit{query-side strategies} improve retrieval by refining input representations, such as rewriting queries through transformation (e.g., rewrite–retrieve–read)\cite{ma2023query}, shifting embeddings toward answer-level semantics with hypothetical document embeddings (HyDE)\cite{gao2023precise}, abstracting queries into higher-level concepts via step-back prompting\cite{zheng2023take}, and combining keyword, semantic, and vector search through hybrid retrieval\cite{karpukhin2020dense}. On \textit{the document side}, methods like ChunkRAG~\cite{singh2024chunkrag} segment text into semantically coherent units and apply multi-stage relevance scoring with self-reflection, critic models, redundancy removal, and dynamic thresholding. \textit{Indexing optimization} further improves dense retrieval by explicitly modeling relevance signals (e.g., source quality, cross-document consistency, citation alignment)~\cite{wu2024retrieval}. 

Despite recent advances, most RAG approaches still process chunks independently, neglecting discourse-level dependencies across and within documents. Ignoring these relationships—such as definitions, coreference links, and discourse transitions—often leads to incomplete or misleading grounding, a problem that is especially severe in multi-document contexts with near-duplicate content and localized references. Addressing this gap requires context-preserving representations and relevance-aware mechanisms that propagate reliability signals across chunks and into generation, ensuring that decisions at time $t$ are conditioned on information established up to $t-1$.

\begin{figure*}[!t]
\centering\resizebox{\textwidth}{!}{%
  \includegraphics[]{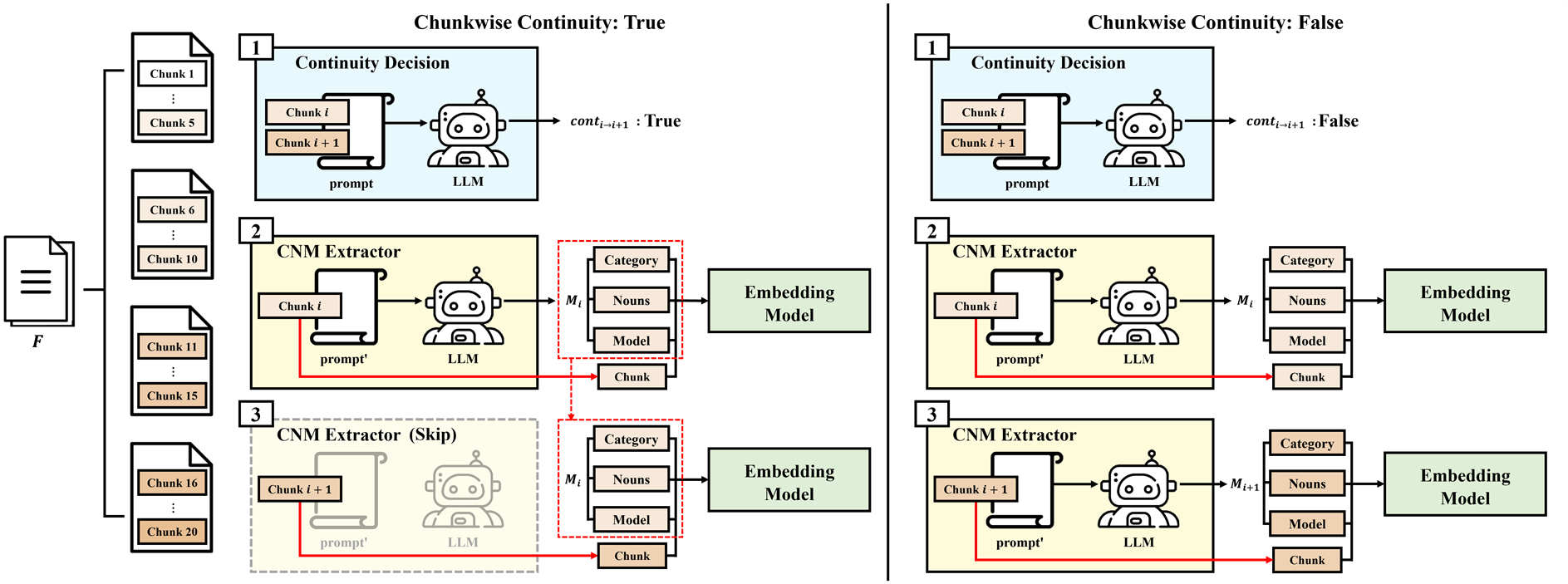}
  }
  \caption{Overview of the CHOP architecture comprising two components: the Continuity Decision module, which determines whether consecutive chunks are continuous, and the CNM-Extractor, which generates compact representations. When continuity holds (left), the next chunk inherits the previous CNM; when it does not (right), a new CNM is extracted. Each chunk is then prefixed with its CNM, and the combined representations are embedded to construct the vector database.}
  \vspace{-3mm}
\end{figure*}

To tackle the limitations of chunk-level independence and semantic conflicts in multi-document collections, we propose \textbf{CHOP}, a Chunkwise Context-Preserving framework for RAG. CHOP enhances retrieval by processing chunks sequentially and enriching them with context-aware embeddings. It is designed for long documents with high lexical or semantic overlap across sub-documents and introduces two key components: (1) the \textit{CNM-Extractor} (Category–Noun–Model Extractor), which generates compact per-chunk signatures capturing categories, key nouns, and model names (2) the \textit{Continuity Decision Module}, which determines whether to inherit the previous chunk’s CNM or extract a new one, depending on contextual continuity. By prefixing each chunk with its CNM, CHOP regularizes the embedding space, mitigates semantic collisions, and reduces retrieval confusion when similar segments coexist—improving relevance in high-overlap settings without retraining the retriever.

\lstdefinestyle{promptstyle}{
    basicstyle=\ttfamily\small,
    breaklines=true,
    columns=fullflexible,
    frame=single,
    backgroundcolor=\color{gray!5},
    keywordstyle=\color{blue!70!black},
    commentstyle=\color{green!40!black},
    stringstyle=\color{red!70!black},
    showstringspaces=false
}

\section{Method}
% CHOP targets long manuals where multiple sub-documents coexist within a single file. 
CHOP is designed for composite long documents characterized by substantial lexical and semantic overlap across sub-documents, a domain that makes retrieval particularly challenging. To handle this, we introduce a sequential pipeline that assigns a consistent per-chunk representation and updates it only when contextual shifts occur. The pipeline consists of two components: (1) the \textbf{CNM-Extractor} (Category–Noun–Model Extractor), which generates key representations from each chunk, and (2) the \textbf{Continuity Decision module}, which determines whether to inherit the previous chunk’s CNM or re-extract a new one based on continuity.

\subsection{CNM-Extractor}
The input file \(F\) is split into a sequence of fixed-size chunks \(\{C_1,\dots,C_n\}\). 
We define CNM as the triplet \(\{\texttt{Category},\texttt{Nouns},\texttt{Model}\}\), which serves as a compact per-chunk signature that anchors each chunk’s semantic position in long documents and facilitates downstream retrieval. The fields are specified as follows:
\begin{itemize}
  \item \textbf{Category}: the broad product family described by the chunk (e.g., \emph{camera}, \emph{air conditioner}, \emph{fan}, \emph{flower}, \emph{boat}).
  \item \textbf{Nouns}: one or two key nouns that capture the core target or action, including parts, operations, or functions (e.g., \emph{air conditioner filter}).
  \item \textbf{Model}: the specific model or series name (e.g., \emph{225B}, \emph{X-SERIES}).
\end{itemize}

As shown in Eq.~\eqref{equation_1}, for each chunk \(C_i\), the CNM-Extractor takes \(C_i\) as input derives its CNM, and produces the corresponding representation \(M_i\).

\begin{equation} 
M_i=\{\texttt{Category}_i,\ \texttt{Nouns}_i,\ \texttt{Model}_i\}
\label{equation_1}
\end{equation}

The CNM-Extractor is an LLM-based module that analyzes a chunk and outputs structured information following a predefined template. Listing~\ref{lst:cnm_extractor} shows an example prompt. The prompt is divided into instructional sections, and any field not explicitly stated or ambiguous must be set to null. The \texttt{Nouns} field requires 1–2 key nouns, with the first strictly constrained to the form “<category> <specific noun>”. The output is enforced as JSON only, avoiding free-form text to prevent format drift. Even when a chunk is processed independently, the extracted CNM is injected as a prefix to stabilize the contextual frame and clarify references to entities, symbols, and units.

% The prompt for CNM extraction is structured into clearly defined instructional sections. Each field must be set to null when it is not explicitly stated in the passage or is ambiguous. The nouns field contains 1–2 key general nouns, with the first item strictly required to be a compound of the form . The output must be JSON only, with no free-form explanation, to prevent format drift. 

% The first chunk is always initialized via the CNM-Extractor:

% \[
% M_1 \leftarrow \mathrm{CNM}(C_1).
% \]

\begin{table*}[t]
\centering
\small
\caption{Comparison of retrieval performance (by Top-$k$). }
\vspace{-3mm}
\label{tab: retrieval-metrics}
\setlength{\tabcolsep}{4.5pt}
\begin{tabular}{l|cccccccccc}
\toprule
% \multirow{2}{*}{\multicolumn{1}{c}{Method}}
\multicolumn{1}{c|}{\multirow{2}{*}{Method}}
& \multicolumn{1}{c}{Top-1}
& \multicolumn{3}{c}{Top-3}
& \multicolumn{3}{c}{Top-5}
& \multicolumn{3}{c}{Top-10} \\
\cmidrule(lr){2-2} \cmidrule(lr){3-5} \cmidrule(lr){6-8} \cmidrule(lr){9-11}
& Hit Rate
& Hit Rate  &  MRR & NDCG
& Hit Rate  & MRR & NDCG
& Hit Rate  & MRR & NDCG \\
\midrule
Native-500T     & 0.8128     & 0.9103 & 0.8551 & 0.8656    & 0.9487 & 0.8637 & 0.8716  &        0.9744 & 0.8676 & 0.8698 \\

Cosine-Chunking  & 0.7077     & 0.8974 & 0.7944 &  0.8309    & 0.9436 & 0.8052 & 0.8442 &       0.9846 & 0.8110 & 0.8389 \\

CHOP    & 0.9077            & 0.9641 & 0.9325 & \textbf{0.9380}       & 0.9821 & 0.9368 & \textbf{0.9380} &        \textbf{0.9923} & \textbf{0.9381} & 0.9291 \\
\bottomrule
\end{tabular}
\vspace{-3mm}
\end{table*}

\subsection{Continuity Decision}

% Continuity Decision $CD(\cdot)$ is an LLM-based decision module that, given a pair of adjacent chunks $(C_i, C_{i+1})$ and a decision prompt, determines whether $C_{i+1}$  $C_i$ or the start of a new document. Eq.~\eqref{equation_2} mathematically defines the continuity decision returned by CD for adjacent chunks.
In multi-document (e.g., manuals, statutes, policy books), topic shifts at chunk boundaries often cause semantic drift or noise amplification, which degrades retrieval quality. The Continuity Detector (CD) addresses this by accurately distinguishing between contextual continuity and heterogeneous topic separation.

The Continuity Decision module $CD(\cdot)$ is an LLM-based classifier that, given a pair of adjacent chunks $(C_i, C_{i+1})$ and a decision prompt, determines whether $C_{i+1}$ is a continuation of the same document as $C_i$ or begins a new one. The decision function is formally defined in Eq.~\eqref{equation_2}.

\begin{equation}
\mathrm{cont}_{i\to i+1} \leftarrow \mathrm{CD}(C_i, C_{i+1}) \in \{\textsf{TRUE},\ \textsf{FALSE}\}.
\label{equation_2}
\end{equation}

%In practice, $CD$ is implemented as an LLM prompted with explicit decision rules (Listing~\ref{lst:manual_segmentation}}.
In practice, $CD$ is implemented as an LLM that is prompted with explicit decision rules (Listing~\ref{lst:manual_segmentation}). The prompt includes several elements to ensure consistent segmentation: a decision goal that instructs the model to treat chunks as part of the same manual unless strong evidence suggests otherwise, a set of decision rules that define explicit conditions for starting a new manual, and paired inputs consisting of the previous anchor text and the current text to enable contextual comparison.

When the continuity decision is TRUE, the two chunks are treated as part of the same topical flow, indicating preserved context. In contrast, FALSE denotes a topic shift and thus a contextual break.

\begin{lstlisting}[style=promptstyle,
caption={Prompt Example for Metadata Extraction},
label={lst:cnm_extractor}]
Extract product CNM from the text below.

Rules:
- category: The most general product family this text describes. (camera, air conditioner, ...). Use null if absent.

- model: The specified model/series(e.g., 225B, X-SERIES). Use null if absent.

- nouns: 1~2 key product nouns.
* The FIRST item MUST be a core compound noun. 
    If the focus is a specific part/task/function within the same category, 
    format it as "<category> <specific noun>"
    * Any remaining item should be a single noun (part/operation/function)
    e.g., battery, charger, filter, nozzle, fan
    
- Always give best estimate + confidence score.
- Return JSON ONLY.

Input Text: 
{text[:1000]}
\end{lstlisting}

Based on the decision, $M_{i+1}$ is determined as follows, as defined in Eq.~\eqref{equation_3}: 
\begin{equation}
M_{i+1} =
\begin{cases}
M_i, & \text{if } \mathrm{cont}_{i\to i+1}=\mathrm{TRUE}\\
\mathrm{CNM}(C_{i+1}), & \text{if } \mathrm{cont}_{i\to i+1}=\mathrm{FALSE}.
\end{cases}
\label{equation_3}
\end{equation}

In segments judged to belong to the same flow, CNM is kept piecewise constant to preserve semantic coherence. In implementation, the CNM of the previous chunk is inherited to maintain contextual continuity. When adjacent chunks are deemed dissimilar, CNM is re-extracted with the CNM-Extractor. Re-extraction explicitly marks a topic shift and prevents contamination or error propagation from prior labels.

Given the decided CNM \(M_i\), we construct \(x_i\) and \(e_i\) as defined in Eq.~\eqref{equation_4}.

\begin{equation}
x_i = [\,\mathrm{PFX}(M_i)\ \Vert\ C_i\,], \qquad \mathbf{e}_i=\mathrm{Embed}(x_i).
\label{equation_4}
\end{equation}

% We embed $x_i$ with \texttt{text-embedding-3-large} to obtain a 3{,}072-dimensional vector $\mathbf{e}_i\in\mathbb{R}^{3072}$. We store the textual input string $x_i$, its vector $\mathbf{e}_i$ in ChromaDB, which uses a Hierarchical Navigable Small World (HNSW) index for approximate nearest-neighbor search.

We embed $x_i$ into a $d$-dimensional vector $\mathbf{e}_i \in \mathbb{R}^d$ and store both the textual input $x_i$ and its embedding $\mathbf{e}_i$ in a vector database. This procedure improves the quality of the evidence set for downstream stages, thereby enhancing the stability and reliability of the RAG pipeline.

\begin{lstlisting}[style=promptstyle,
    caption={Prompt Example for Continuity Decision},
    label={lst:manual_segmentation}]
Decision goal: Be conservative and keep texts in the SAME manual ...
Decision rules:
- same=true when: same product category/model, or just section shift...
- same=false when: product category changes, new model (e.g., 225B->226R), or new doc markers...
Previous anchor:
Current text to judge:
\end{lstlisting}
\vspace{-3mm}

% At test time, a query $q$ is embedded with the same model as $\mathbf{e}_q$. We compute cosine similarity to each stored vector:
% \begin{equation}
% \mathrm{sim}(\mathbf{e}_q,\mathbf{e}_i)
% = \frac{\mathbf{e}_q \cdot \mathbf{e}_i}{\lVert \mathbf{e}_q \rVert\, \lVert \mathbf{e}_i \rVert}.
% \end{equation}
% Finally, we select the top-$k$ texts with the highest similarity as relevant documents, which are then used to construct prompts for the downstream generator. By encoding the CNM prefix together with the original chunk, the embedding space is regularized, semantic collisions among similar chunk are mitigated, and both accuracy and evidence interpretability are improved in the subsequent generation stage.

% \begin{lstlisting}[float, 
% style=promptstyle, 
% caption={Prompt Example for Metadata Extraction}
% label = {lst:metadata_extraction},
% placement = H]
% Extract product metadata from the text below.

% Rules:
% - category: General product family (camera, air conditioner, ...). Use null if absent.
% - model: Specified model/series (225B, ...). Use null if absent.
% - nouns: 1–2 key product nouns.
%   * First item = core compound noun or "<category> <specific noun>" (camera battery, ...).
%   * Second item = single noun (filter, safety, ...).
% - Always give best estimate + confidence score.
% - Return JSON ONLY.

% {text[:1000]}
% \end{lstlisting}

\begin{table*}[t]
\centering
\small
\caption{Comparison of generation performance (by Top-$k$).}
\vspace{-3mm}
\label{tab: generation-metrics}
\setlength{\tabcolsep}{4.5pt}
\begin{adjustbox}{max width=\textwidth}
\begin{tabular}{l|cccccccccccc}
\toprule
% \multirow{2}{*}{\multicolumn{1}{c}{Method}}
\multicolumn{1}{c|}{\multirow{2}{*}{Method}}
& \multicolumn{3}{c}{Top-1}
& \multicolumn{3}{c}{Top-3}
& \multicolumn{3}{c}{Top-5}
& \multicolumn{3}{c}{Top-10} \\
\cmidrule(lr){2-4} \cmidrule(lr){5-7} \cmidrule(lr){8-10} \cmidrule(lr){11-13}
&  F1 & ROUGE-L  & BERTScore
&  F1 & ROUGE-L  & BERTScore
&  F1 & ROUGE-L  & BERTScore
&  F1 & ROUGE-L  & BERTScore \\
\midrule
Native-500T     & 0.2763 & 0.2472 &  0.6992    & 0.3497 & 0.3144 & 0.7253     & 0.3792 & 0.3394 &  0.7338    & 0.4072 &0.3573 & 0.7381 \\

Cosine-Chunking  & 0.2399 & 0.2143 & 0.6849     & 0.3093 & 0.2730 & 0.7119     & 0.3351 & 0.2962 &  0.7206 &     0.3577  & 0.3107 & 0.7256 \\

CHOP    &  0.2760 & 0.2475 & 0.6998        &  0.3513 &  0.3164 & 0.7257     & 0.3814 & 0.3412  & 0.7349 &       \textbf{0.4080} & \textbf{0.3591} &  \textbf{0.7396} \\
\bottomrule
\end{tabular}
\end{adjustbox}
\vspace{-3mm}
\end{table*}

\section{Experiments}
\subsection{Implementation Details}
All chunks and queries are encoded with OpenAI’s \texttt{text-embedding\-3-large}\cite{openai_textembedding3_large}, producing 3{,}072-dimensional vectors $\mathbf{e}_i \in \mathbb{R}^{3072}$. The embeddings are stored in ChromaDB\cite{chroma2023}, with an HNSW (Hierarchical Navigable Small World) index~\cite{malkov2018efficient} used for approximate nearest-neighbor search. We perform retrieval by computing cosine similarity between the query and chunk embeddings, and the top-$k$ most similar chunks are selected as matches. Because the focus of this study is classification and decision-making rather than text generation, the temperature is fixed at $0$ to ensure deterministic outputs and reproducibility. The backbone LLM in all experiments is Gemma-12B~\cite{team2024gemma}.

\textbf{Dataset.} We evaluate on MRAMG-Bench~\cite{yu2025mramg}, a benchmark derived from product manuals in ManualsLib~\cite{manualslib}. While the original dataset presents each manual as multiple split lists or fragments, we reconstruct each manual into a single continuous file to better emulate real-world usage. This weakens explicit document-boundary cues, requiring retrieval to rely more heavily on contextual continuity. At the same time, natural topic transitions and inter-section dependencies are preserved, making the corpus more faithful to real manual structures. All methods are evaluated under this reconstructed single-document setting.

% 실험 디테일만 넣기 

\subsection{Retrieval Evaluation} \label{3.3}
% This study evaluates the effect of CHOP’s consistent prefix injection while isolating only the variation in chunk composition. The retrieval pipeline (embedding, indexing, and similarity computation) is held constant, and we assess how alternative chunking strategies impact retrieval performance.
This study systematically verifies whether the proposed method, CHOP, improves retrieval performance. To isolate the effect of CHOP’s consistent prefix injection, the retrieval stack—embedding, indexing, and similarity computation—is kept fixed, and only the chunk composition is varied. We then apply diverse chunking strategies and quantify their impact on retrieval performance.

\subsubsection{Baseline}
\begin{itemize}
\item \textbf{Naive-500T:} Each document is uniformly split into fixed-length chunks of 500 tokens with an overlap of 100 tokens.
\item \textbf{Cosine-Chunking~\cite{singh2024chunkrag}:} Each document is split into sentences, and topic-shift boundaries are detected using a cosine similarity threshold of 0.35 between consecutive sentences, producing adaptively sized chunks.
\end{itemize}

\subsubsection{Evaluation Metrics} 
Table~\ref{tab: retrieval-metrics}  compares the retrieved set for each query against its gold evidence. Under a Top-K assumption, we report standard retrieval metrics: Hit Rate at K (Hit@K), Mean Reciprocal Rank at K (MRR@K), and Normalized Discounted Cumulative Gain at K (NDCG@K~\cite{jarvelin2002cumulated}). Hit@K measures the proportion of cases in which the Top-K retrieved set contains at least one gold-evidence chunk. MRR@$K$ computes the average reciprocal rank of the first relevant chunk per query within the Top-$K$ list. NDCG@$K$ applies logarithmic position discounting to Discounted Cumulative Gain (DCG) and normalizes it by the ideal DCG (IDCG), capturing overall ranking quality across the Top-$K$.

\subsubsection{Experiment Result}
CHOP consistently achieves the best performance across all $K$. Specifically, its Top-1 Hit Rate was 0.9077, outperforming Native-500T (0.8128) and Cosine-Chunking (0.7077). As $K$ increases, the Hit Rates of all three methods converge to the 0.97–0.99 range, yet CHOP’s maintains clear advantages in MRR@$K$ and NDCG@$K$, which reflect ranking quality. This indicates that CHOP’s key strength lies not only in retrieving the correct evidence but also in ranking it higher.
The performance gains can be attributed to two main factors.
First, CHOP preserves essential contextual information during chunk segmentation, thereby strengthening semantic alignment with queries.
Second, by identifying semantically related chunks and normalizing them under the same CNM, it consolidates related evidence into coherent semantic units, enabling more consistent retrieval and ranking.
As a result, CHOP enhances ranking fidelity, a critical aspect of retrieval quality, and mitigates error propagation in the generation stage.

\subsection{Generation Evaluation}
Table~\ref{tab: generation-metrics} presents the detailed performance of responses generated from retrieved evidence. This experiment evaluates the final generation quality of the RAG pipeline—specifically, whether retrieval via CHOP enables accurate, hallucination-free generation. We use the QA subset of MRAMG-Bench as the evaluation dataset, and the baseline configuration follows Section~\ref{3.3}. For each method, retrieved results serve as input to the LLM, which generates answers that are then evaluated against gold references.

\subsubsection{Evaluation Metrics}
\begin{itemize}
\item \textbf{F1:} Computes the harmonic mean of precision and recall from the token overlap between prediction and reference, measuring how exactly the same words are recovered.
\item \textbf{ROUGE-L:} Evaluates order-preserving similarity based on the length of the longest common subsequence between prediction and reference.
\item \textbf{BERTScore~\cite{zhang2019bertscore}:} Uses pretrained embeddings to match token-level semantic similarity, producing a score that is robust to paraphrases and lexical variation and thus captures semantic equivalence well.
\end{itemize}

\subsubsection{Experiment Result}
Overall, all metrics increase monotonically as $K$ grows, and CHOP consistently exhibits superior performance than Naive-500T across nearly all $K$. In particular, for Top-3/5/10, CHOP yields steady absolute gains in F1 and ROUGE-L (+0.0266 to +0.0753), with additional improvements in BERTScore. By contrast, Cosine-Chunking records the lowest performance across all metrics and values of $K$. These results show that CHOP’s more accurate retrieval consistently translates into higher generation quality than Naive-500T. Although the absolute improvements are modest, they are consistent across metrics and $K$, indicating that CHOP suppresses irrelevant retrievals and prioritizes useful ones, thereby reducing errors propagated to the generation stage. While larger $K$ improves quality, the marginal gains diminish; thus, the range $K=5$–10 provides a practical operating point. Within the same budget, CHOP achieves higher generation quality than Naive-500T, demonstrating better budget efficiency.

% \begin{tcolorbox}[breakable=true, colback=gray!5!white, colframe=black!40, sharp corners,
% title={Example of CHOP Generation Prompt},space to upper] 
    
% \textbf{<s>[INST]<<SYS>>}

% \end{tcolorbox}

\lstdefinestyle{promptstyle}{
    basicstyle=\ttfamily\small,
    breaklines=true,
    columns=fullflexible,
    frame=single,
    backgroundcolor=\color{gray!5},
    keywordstyle=\color{blue!70!black},
    commentstyle=\color{green!40!black},
    stringstyle=\color{red!70!black},
    showstringspaces=false
}

\section{Conclusion}

This paper presents CHOP, a method that improves retrieval in RAG pipelines by placing correct evidence more frequently and at higher ranks. CHOP (i) standardizes category/noun/model information through prefixes to reduce embedding collisions and (ii) preserves cross-segment continuity with continuity-aware chunking, maintaining semantic consistency and mitigating topic drift in long, stitched technical documents. Empirically, CHOP consistently improves retrieval and generation metrics, achieving up to +7.53\% in NDCG@10. It promotes not only inclusion of supporting evidence but also its higher placement, while balancing efficiency and effectiveness under limited context budgets. Overall, CHOP offers a practical baseline for building efficient and reliable RAG systems in real-world settings. For future work, we plan to explore adaptive prefixing and dynamic continuity modeling to effectively handle evolving domain knowledge and streaming inputs. Furthermore, we aim to develop lightweight strategies and efficient inference techniques to reduce the computational and time costs associated with using large language models.

% \subsection{Complexity}
% The procedure runs in a single forward pass over chunks: one extraction for \(C_1\), \(n-1\) continuity decisions with inherit-or-reextract updates, and $n$ prefix+embedding steps, yielding overall \(O(n)\) time.

% \begin{acks}
% Thanks to Corporate Lorem - \url{https://corporatelorem.kovah.de/} and ``lipsum.com'' - \url{https://www.lipsum.com/} for the ``Lorem Ipsum'' text.
% \end{acks}

\bibliographystyle{ACM-Reference-Format}
\bibliography{main}

\end{document}